\documentclass[10pt,twocolumn,letterpaper]{article}

\usepackage{cvpr}              

\usepackage{graphicx}
\usepackage{amsmath}
\usepackage{amssymb}
\usepackage{booktabs}

\usepackage[accsupp]{axessibility}

\usepackage{xcolor} 
\usepackage{times}
\usepackage{epsfig}
\usepackage{graphicx}
\usepackage{amsmath}
\usepackage{amssymb}
\usepackage[bb=dsserif]{mathalpha}
\usepackage{bm}

\usepackage{booktabs,colortbl,tabularx}
\usepackage{pifont}%

\usepackage{mathtools}
\usepackage{commath}
\usepackage{algorithm}
\usepackage[noend]{algpseudocode}

\usepackage{multirow}
\usepackage{comment} 
\usepackage{enumitem}

\definecolor{Gray}{gray}{0.9}

\newcommand{\cmark}{\ding{51}}%
\newcommand{\xmark}{\ding{55}}%

\definecolor{battleshipgrey}{rgb}{0.52, 0.52, 0.51}

\usepackage[pagebackref,breaklinks,colorlinks]{hyperref}

\usepackage[capitalize]{cleveref}
\crefname{section}{Sec.}{Secs.}
\Crefname{section}{Section}{Sections}
\Crefname{table}{Table}{Tables}
\crefname{table}{Tab.}{Tabs.}

\def\confName{CVPR}
\def\confYear{2023}

\begin{document}

\title{AV-SAM: Segment Anything Model Meets Audio-Visual \\ Localization and Segmentation}

\author{%
  Shentong Mo\\
  Carnegie Mellon University
  \and
  Yapeng Tian\thanks{Corresponding author.} \\
  University of Texas at Dallas 
}

\maketitle

\begin{abstract}

Segment Anything Model (SAM) has recently shown its powerful effectiveness in visual segmentation tasks.
However, there is less exploration concerning how SAM works on audio-visual tasks, such as visual sound localization and segmentation.
In this work, we propose a simple yet effective audio-visual localization and segmentation framework based on the Segment Anything Model, namely AV-SAM, that can generate sounding object masks corresponding to the audio.
Specifically, our AV-SAM simply leverages pixel-wise audio-visual fusion across audio features and visual features from the pre-trained image encoder in SAM to aggregate cross-modal representations.
Then, the aggregated cross-modal features are fed into the prompt encoder and mask decoder to generate the final audio-visual segmentation masks.
We conduct extensive experiments on Flickr-SoundNet and AVSBench datasets. 
The results demonstrate that the proposed AV-SAM can achieve competitive performance on sounding object localization and segmentation.

\end{abstract}


\section{Introduction}

When we hear a lion roaring in the zoo, we are naturally aware of where the lion is sitting due to the strong correlation between audio signals and sounding objects in our world.
This intelligent perception from humans attracts many researchers to explore joint audio-visual learning for visual sound localization and segmentation.

Audio-visual localization and segmentation is a challenging problem that predicts pixel-wise masks of sounding objects in a video.
To tackle this task, recent methods~\cite{mo2022EZVSL,mo2022SLAVC} manually designed diverse contrastive learning pipelines with audio-visual alignment as clues to localize sound sources.
More recently, AVGN~\cite{mo2023audiovisual} leveraged cross-modal grouping with class-wise semantics to predict each source from the audio mixture.
In this work, we will solve the problem by extracting compact features with pixel-wise audio-visual fusion based on Segment Anything Model.

Segment Anything Model (SAM)~\cite{kirillov2023segany} has recently shown its impressive effectiveness in image segmentation tasks.
This promptable foundation model was trained on 1B masks from 11M images and composed of three main components: an image encoder, a prompt encoder, and a mask decoder.
However, no work is yet to explore how SAM works on challenging audio-visual tasks, including visual sound localization and segmentation.

The main challenge is that audio is not naturally aligned with all objects that exist in the video. 
This motivates us to learn audio-aligned visual features for each mask prompt from the video to guide mask generation in SAM. 
To address the problem, our key idea is to leverage pixel-wise audio-visual fusion across audio and visual features to aggregate cross-modal representations as the input to the audio-visual mask decoder, which is simpler than previous audio-visual localization and segmentation methods.

To this end, we propose a novel Segment Anything Model with pixel-wise Audio-Visual fusion, namely AV-SAM, that can predict sounding object masks corresponding to the audio signal.
Specifically, our AV-SAM aggregates cross-modal representations across audio representations and visual features from the pre-trained image encoder in SAM by using an explicit pixel-wise audio-visual fusion.
Then, the audio-visual mask decoder takes aggregated cross-modal features and prompt embeddings as input to generate the final segmentation masks.

Extensive experiments are conducted on Flickr-SoundNet and AVSBench datasets. 
The empirical results demonstrate that the proposed AV-SAM can achieve competitive performance against SAM on sounding object localization and segmentation.
In addition, qualitative mask visualizations vividly showcase the advantage of our AV-SAM in audio-visual localization and segmentation.

\begin{figure*}[t]
    \centering
    \includegraphics[width=0.85\linewidth]{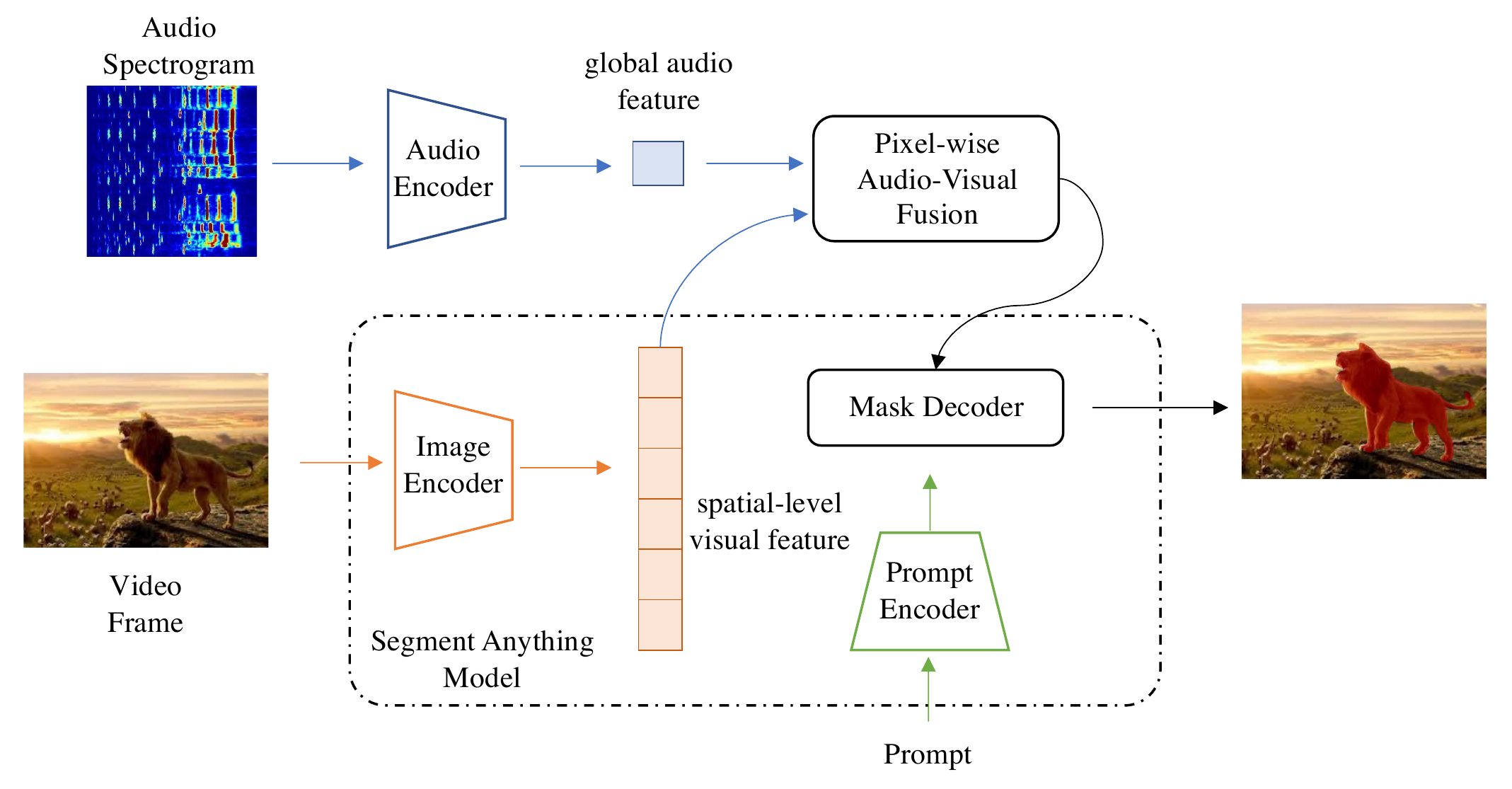}
    \vspace{-0.5em}
    \caption{Illustration of the proposed Segment Anything Model for Audio-Visual
localization and segmentation (AV-SAM).
The pixel-wise audio-visual fusion module aggregates cross-modal representations across audio features and visual features from the pre-trained audio and image encoder.
Then, aggregated cross-modal features and prompt embeddings from the prompt encoder are updated into the mask decoder to generate the final segmentation mask.
    }
    \label{fig: main_img}
    \vspace{-1em}
\end{figure*}

\section{Related Work}

\noindent{\textbf{Audio-Visual Joint Learning.}}
Audio-visual joint learning has been addressed in many previous works to learn the audio-visual alignment between two distinct modalities from videos for many audio-visual tasks, such as audio-visual spatialization~\cite{Morgado2018selfsupervised,Morgado2020learning}, audio-event localization~\cite{tian2018ave} and audio-visual parsing~\cite{tian2020avvp,mo2022multimodal}.
In this work, our main focus is to learn discriminative cross-modal representations for audio-visual localization and segmentation, which is more challenging than the tasks discussed above.

\noindent{\textbf{Audio-Visual Localization \& Segmentation.}}
Audio-visual localization and segmentation is a typical and challenging problem that predicts the location of individual sound sources in a video in a heatmap or mask manner.
In recent years, researchers~~\cite{mo2022EZVSL,mo2022SLAVC,mo2023audiovisual,zhou2022avs} have explored diverse and manually-designed pipelines to learn the audio-visual correspondence for localizing single- or multiple-source sounds.
Different from them, we propose a simple framework based on the recent visual foundation model, Segment Anything Model, to achieve audio-visual localization and segmentation simultaneously.

\noindent{\textbf{Segment Anything Model.}}
Segment Anything Model (SAM)~\cite{kirillov2023segany} is a promptable model trained on one billion masks from 11M images for image segmentation.
This visual foundation model consists of three main components: an image encoder, a prompt encoder, and a mask decoder.
Although the model showcased impressive visual segmentation results, how SAM performs on audio-visual localization and segmentation is unknown. 
In contrast, we design a novel framework based on SAM with pixel-wise audio-visual fusion for aggregating cross-modal representations between audio and visual modalities.

\vspace{-0.5em}

\section{Method}

Given an image and audio, our goal is to predict pixel-wise masks of sounding objects on the image. 
We propose a novel Segment Anything Model with Audio-Visual fusion named AV-SAM for visual sound localization and segmentation, which mainly consists of two modules, Pixel-wise Audio-Visual Fusion~\ref{sec:pavf} and Audio-Visual Mask Decoder in Section~\ref{sec:avmd}.

\subsection{Preliminaries}

In this section, we first describe the problem setup and notations, and then revisit the mask generation process in SAM~\cite{kirillov2023segany} for image segmentation.

\noindent\textbf{Problem Setup and Notations.}
Given a spectrogram of audio and an image, our target is to predict the spatial location of individual sound sources in the image.
Let $\mathcal{D} = {(a_i, v_i) : i = 1,...,N}$ be a dataset of audio $a_i\in\mathbb{R}^{T\times F}$ and visual $v_i\in\mathbb{R}^{3\times H\times W}$ pairs.
Note that $T, F$ denotes the time and frequency dimension of the audio spectrogram, respectively
Following previous work~\cite{zhou2022avs,mo2022EZVSL}, we first encode the audio and visual inputs using a dual-stream encoder and projection heads, denoted as $f_a(\cdot), g_a(\cdot)$ and $f_v(\cdot), g_v(\cdot)$ for the audio and images, separately. 
The audio encoder calculates global audio features $\mathbf{a}_i = g_a(f_a(a_i)), \mathbf{a}_i\in\mathbb{R}^{1\times D}$
and the visual encoder generates multi-scale spatial-level features $\{\mathbf{v}_i^{s}\}_{s=1}^S = g_v(f_v(v_i)), \mathbf{v}_i^{s}\in\mathbb{R}^{D\times H^s\times W^s}$ for each $s$th stage.

\noindent\textbf{Revisit SAM.}
To address the visual segmentation problem, SAM~\cite{kirillov2023segany} introduced a prompt encoder $f_p(\cdot)$ and a lightweight mask decoder $f_m(\cdot)$ to generate the segmentation mask.
Specifically, the prompt encoder took as input positional embeddings with sparse prompt types, such as points, boxes, and text, and dense mask prompt embeddings based on convolutions.
Then the mask decoder applied a transformer decoder block with self-/cross- attention and a dynamic mask prediction head to predict the final mask.

\subsection{Pixel-wise Audio-Visual Fusion}\label{sec:pavf}

To solve the audio-visual segmentation problem, we are inspired by AVS~\cite{zhou2022avs} and introduce the pixel-wise audio-visual fusion module to encode the multi-scale spatial-level visual features and global audio representations to update the input to the following mask decoder.
After the cross-modal fusion, the audio-visual features $\mathbf{z}_i^{s}$ at the $s$th stage is updated as:
\begin{equation}\label{eq:avf}
    \mathbf{z}_i^{s} = \mathbf{v}_i^{s} + \mu\left(\dfrac{\theta(\mathbf{v}_i^{s})\phi(\hat{\mathbf{a}}_i)^\top}{H^sW^s} \omega(\mathbf{v}_i^{s})\right)
\end{equation}
where $\hat{\mathbf{a}}_i\in\mathbb{R}^{D\times H^s\times W^s}$ denotes the duplicated version of the global audio representation $\mathbf{a}_i$ that repeats $H^sW^s$ times at the $s$th stage.
Here, $\mu(\cdot),\theta(\cdot),\phi(\cdot),\omega(\cdot)$ denote the $1\times1\times1$ convolution operator.
With this explicit audio-visual fusion, we will push the learned visual token embeddings to be discriminatively aligned with global audio features.

\subsection{Audio-visual Mask Decoder}\label{sec:avmd}

With the benefit of pixel-wise audio-visual fusion, we update the original visual features from pre-trained image encoder in SAM~\cite{kirillov2023segany} with the last stage of multi-scale feature maps $\mathbf{z}_i^{s}$.
Then those updated multi-stage feature maps are passed into the mask decoder and prompt encoder in SAM~\cite{kirillov2023segany} to generate the final output mask $\mathbf{M}\in\mathbb{R}^{H\times W}$.
With the pixel-level annotation $\mathbf{Y}$ as supervision, we simply apply the binary cross entropy (BCE) loss between the prediction and label as:
\begin{equation}
    \mathcal{L} = \mbox{BCE}(\mathbf{M}, \mathbf{Y})
\end{equation}
During inference, we follow the prior work~\cite{kirillov2023segany} and predict the final audio-visual segmentation mask for evaluation.

\section{Experiments}

\subsection{Experimental setup}

\noindent \textbf{Datasets.}
We use a subset of 144k pairs in VGG-Sound~\cite{chen2020vggsound} for training, and test the model on Flickr-SoundNet testset~\cite{Senocak2018learning} with 250 audio-visual pairs of sounding objects.
Note that we use an ImageNet~\cite{imagenet_cvpr09} pre-trained ResNet50~\cite{he2016resnet} to generate the pseudo mask by bilinear interpolation of the feature map. 
AVSBench~\cite{zhou2022avs} includes 4,932 videos (in total 10,852 frames) from 23 categories including instruments, humans, animals, etc.
Following prior work~\cite{zhou2022avs}, we use the same split of 3,452/740/740 videos for train/val/test.

\noindent \textbf{Evaluation Metrics.}
For visual sound localization, we follow the previous method~\cite{mo2022EZVSL,mo2023audiovisual} and use the average precision at the pixel-wise average precision (AP), Intersection over Union (IoU), and Area Under Curve (AUC).
For audio-visual segmentation, we apply mIoU and F1 scores as evaluation metrics, following the previous work~\cite{zhou2022avs}.

\noindent \textbf{Implementation.}
For input visual frames, the resolution is resized to $1024 \times 1024$. 
For input audio, we use the log spectrograms with the length of $3s$ at a sampling rate of $22050$Hz. 
Following the prior work~\cite{mo2022EZVSL}, we apply STFT to generate an input tensor of shape $257 \times 300$ (\textit{i.e.}, $257$ frequency bands over $300$ timesteps) by using 50ms windows with a 25ms hop. 
We use the lightweight ResNet18~\cite{he2016resnet} as the audio encoder, and initialize the visual model using weights released from SAM~\cite{kirillov2023segany}.
The model is trained for 100 epochs using a batch size of 128 and the Adam optimizer with a learning rate of $1e-4$.

\begin{table}[t]
	\renewcommand\tabcolsep{6.0pt}
	\centering
	\scalebox{0.85}{
		\begin{tabular}{l|ccc|cc}
			\toprule
			\multirow{2}{*}{Method} & \multicolumn{3}{c|}{Flickr-SoundNet} & \multicolumn{2}{c}{AVS-Bench}  \\
			& AP(\%) & IoU(\%) & AUC(\%) & mIoU(\%)  & F1(\%) \\ 	
			\midrule
			SAM~\cite{kirillov2023segany} &20.02 & 41.60 & 20.46 & 29.67 & 42.06 \\
                AV-SAM & \textbf{28.93} & \textbf{50.80} & \textbf{29.76} & \textbf{40.47} & \textbf{56.57} \\
			\bottomrule
			\end{tabular}}
   \vspace{-0.5em}
   \caption{Quantitative results of single-source localization and segmentation on Flickr-SoundNet and AVS-Bench datasets.}
   \label{tab: exp_sota}
   \vspace{-0.5em}
\end{table}

\begin{table}[t]
	\renewcommand\tabcolsep{6.0pt}
	\centering
	\scalebox{0.85}{
		\begin{tabular}{ccc|cc}
			\toprule
			Mask & Prompt & Image &  \multirow{2}{*}{mIoU(\%)}  & \multirow{2}{*}{F1(\%)} \\
            Decoder & Encoder & Encoder & & \\
			\midrule
			\xmark & \xmark & \xmark  & 32.58 & 45.36 \\
                \cmark & \xmark & \xmark  & 34.11 & 49.67 \\
                \cmark & \cmark & \xmark  & 36.63 & 51.58 \\
                \cmark & \cmark & \cmark  & \textbf{40.47} & \textbf{56.57} \\
			\bottomrule
			\end{tabular}}
   \vspace{-0.5em}
   \caption{Exploration study on the frozen (\xmark) and fine-tuning (\cmark) in modules of SAM for audio-visual localization and segmentation.}
   \label{tab: ab_module}
   \vspace{-1.5em}
\end{table}

\subsection{Comparison to prior work}

In this work, we propose a novel and effective framework based on SAM~\cite{kirillov2023segany} for audio-visual localization and segmentation. 
In order to validate the effectiveness of the proposed AV-SAM, we comprehensively compare it to the visual foundation baseline.

For visual sound localization on Flickr-SoundNet, we report the quantitative comparison results in Table~\ref{tab: exp_sota}.
As can be seen, we achieve the best results in terms of all metrics for two benchmarks compared to SAM~\cite{kirillov2023segany}, the competitive visual foundation baseline.
In particular, the proposed AV-SAM advantageously outperforms SAM~\cite{kirillov2023segany} by 8.91 AP, 9.20 IoU, and 9.30 AUC on the Flick-SoundNet dataset.
Moreover, we achieve superior results gains of 10.80 mIoU and 14.51 F1, which indicates the importance of pixel-wise audio-visual fusion to aggregate cross-modal inputs as the guidance for the mask decoder to generate discriminative segmentation masks.
These significant improvements demonstrate the superiority of our method against SAM~\cite{kirillov2023segany} in audio-visual localization and segmentation.

\begin{figure}[t]
\centering
\includegraphics[width=0.99\linewidth]{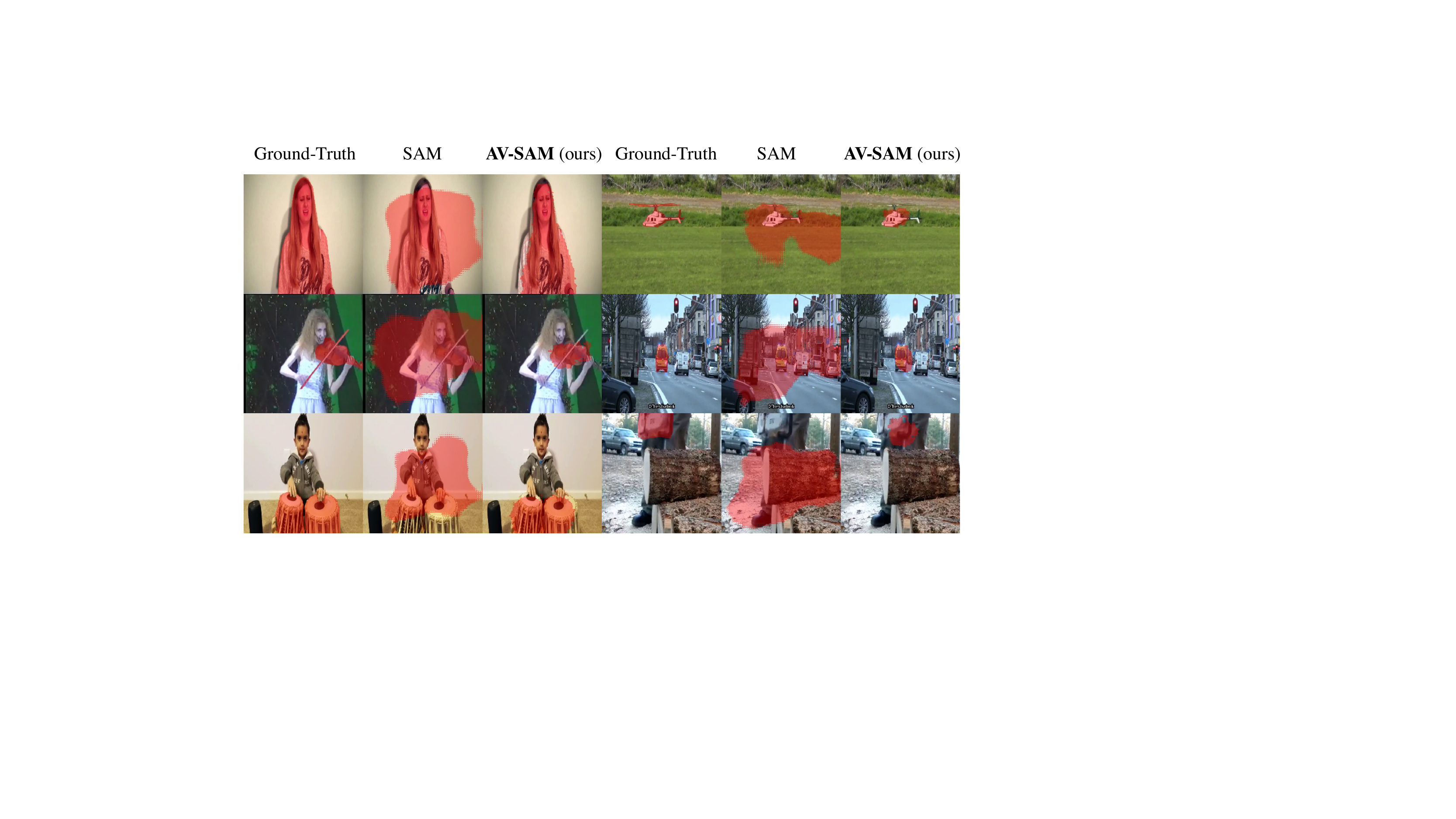}
\vspace{-0.5em}
\caption{Qualitative comparisons with SAM on audio-visual segmentation. 
The proposed AV-SAM produces more accurate and high-quality segmentation masks for sounding sources. 
}
\label{fig: vis_source}
\vspace{-1.5em}
\end{figure}

In order to qualitatively evaluate the segmentation masks, we compare the proposed AV-SAM with SAM~\cite{kirillov2023segany} on AVSBench in Figure~\ref{fig: vis_source}.
Note that we only visualize one mask merged from SAM~\cite{kirillov2023segany} with high confidence scores.
From comparisons, we can observe that without explicit audio-visual fusion, SAM~\cite{kirillov2023segany}, the strong visual segmentation baseline, performs worse on sounding objects.
For example, given an image of a girl playing violin, the baseline model tends to predict the mask across both the girl and the violin. 
In contrast, the quality of segmentation masks generated by our method is much better and closer to the ground truth.
These visualizations further showcase the superiority of our simple AV-SAM in visual sound segmentation with explicit pixel-wise audio-visual fusion based on SAM~\cite{kirillov2023segany}.

\subsection{Experimental analysis}

In this section, we performed ablation studies to demonstrate the effect of frozen and fine-tuning pre-trained weights from SAM~\cite{kirillov2023segany}.
To explore such an effect more comprehensively, we ablate the necessity of fine-tuning parameters from each module (mask decoder, prompt encoder, image encoder) in Table~\ref{tab: ab_module}.
We can observe that fine-tuning the mask decoder based on the vanilla baseline increases the results of audio-visual segmentation (by 1.53 mIoU and 4.31 F1), which shows the benefit of audio-visual mask decoder in generating accurate masks from aggregated cross-modal features.
Meanwhile, fine-tuning the prompt encoder also increases the visual sound source segmentation performance in terms of all metrics.
More importantly, fine-tuning all three modules together in the baseline highly raises the performance by 7.89 mIoU and 11.21 F1.
These improving results validate the importance of fine-tuning pre-trained weights from SAM in extracting cross-modal semantics from audio and images in audio-visual segmentation.

\section{Conclusion}

In this work, we present AV-SAM, a simple yet effective audio-visual framework based on the SAM, namely AV-SAM, that can generate sounding object masks in the image.
We introduce pixel-wise audio-visual fusion to aggregate cross-modal representations between audio and visual features in SAM.
Then, we leverage aggregated cross-modal features to generate the final audio-visual segmentation masks using the prompt encoder and mask decoder.
Empirical experiments on Flickr-SoundNet and AVSBench datasets demonstrate the competitive performance of the proposed AV-SAM on visual sound localization and segmentation.

{\small
\bibliographystyle{ieee_fullname}
\bibliography{reference}

\begin{thebibliography}{10}\itemsep=-1pt

\bibitem{chen2020vggsound}
Honglie Chen, Weidi Xie, Andrea Vedaldi, and Andrew Zisserman.
\newblock Vggsound: A large-scale audio-visual dataset.
\newblock In {\em ICASSP}, 2020.

\bibitem{imagenet_cvpr09}
Jia Deng, Wei Dong, Richard Socher, Li-Jia. Li, Kai Li, and Li Fei-Fei.
\newblock {ImageNet: A Large-Scale Hierarchical Image Database}.
\newblock In {\em CVPR}, 2009.

\bibitem{he2016resnet}
Kaiming He, Xiangyu Zhang, Shaoqing Ren, and Jian Sun.
\newblock Deep residual learning for image recognition.
\newblock In {\em CVPR}, pages 770--778.

\bibitem{kirillov2023segany}
Alexander Kirillov, Eric Mintun, Nikhila Ravi, Hanzi Mao, Chloe Rolland, Laura
  Gustafson, Tete Xiao, Spencer Whitehead, Alexander~C. Berg, Wan-Yen Lo, Piotr
  Doll{\'a}r, and Ross Girshick.
\newblock Segment anything.
\newblock {\em arXiv:2304.02643}, 2023.

\bibitem{mo2022SLAVC}
Shentong Mo and Pedro Morgado.
\newblock A closer look at weakly-supervised audio-visual source localization.
\newblock In {\em NeurIPS}, 2022.

\bibitem{mo2022EZVSL}
Shentong Mo and Pedro Morgado.
\newblock Localizing visual sounds the easy way.
\newblock In {\em ECCV}, page 218–234, 2022.

\bibitem{mo2022multimodal}
Shentong Mo and Yapeng Tian.
\newblock Multi-modal grouping network for weakly-supervised audio-visual video
  parsing.
\newblock In {\em NeurIPS}, 2022.

\bibitem{mo2023audiovisual}
Shentong Mo and Yapeng Tian.
\newblock Audio-visual grouping network for sound localization from mixtures.
\newblock {\em arXiv preprint arXiv:2303.17056}, 2023.

\bibitem{Morgado2020learning}
Pedro Morgado, Yi Li, and Nuno Nvasconcelos.
\newblock Learning representations from audio-visual spatial alignment.
\newblock In {\em NeurIPS}, pages 4733--4744, 2020.

\bibitem{Morgado2018selfsupervised}
Pedro Morgado, Nuno Nvasconcelos, Timothy Langlois, and Oliver Wang.
\newblock Self-supervised generation of spatial audio for 360\textdegree video.
\newblock In {\em NeurIPS}, 2018.

\bibitem{Senocak2018learning}
Arda Senocak, Tae-Hyun Oh, Junsik Kim, Ming-Hsuan Yang, and In~So Kweon.
\newblock Learning to localize sound source in visual scenes.
\newblock In {\em CVPR}, 2018.

\bibitem{tian2020avvp}
Yapeng Tian, Dingzeyu Li, and Chenliang Xu.
\newblock Unified multisensory perception: Weakly-supervised audio-visual video
  parsing.
\newblock In {\em ECCV}, page 436–454, 2020.

\bibitem{tian2018ave}
Yapeng Tian, Jing Shi, Bochen Li, Zhiyao Duan, and Chenliang Xu.
\newblock Audio-visual event localization in unconstrained videos.
\newblock In {\em ECCV}, 2018.

\bibitem{zhou2022avs}
Jinxing Zhou, Jianyuan Wang, Jiayi Zhang, Weixuan Sun, Jing Zhang, Stan
  Birchfield, Dan Guo, Lingpeng Kong, Meng Wang, and Yiran Zhong.
\newblock Audio-visual segmentation.
\newblock In {\em ECCV}, 2022.

\end{thebibliography}
}

\end{document}